\pdfoutput=1

\documentclass[11pt]{article}

\usepackage[final]{acl}

\usepackage{times}
\usepackage{latexsym}
\usepackage{amsmath}
\usepackage[capitalise]{cleveref}
\usepackage{multirow}
\usepackage{tcolorbox}
\usepackage{float}
\usepackage[T1]{fontenc}

\usepackage[utf8]{inputenc}

\usepackage{microtype}
\usepackage{booktabs}

\usepackage{inconsolata}
\usepackage{xspace}

\usepackage{graphicx}

\definecolor{pli-color}{HTML}{FF6A6A}

\title{Vision Language Model Helps Private Information De-Identification in Vision Data}

\author{
{\bf Tiejin Chen}$^{1}$, {\bf Pingzhi Li}$^{2}$, {\bf Kaixiong Zhou}$^{3}$, {\bf Tianlong Chen}$^{2}$, {\bf Hua Wei}$^{1}$ \\
$^{1}$Arizona State University \\
$^{2}$University of North Carolina at Chapel Hill \\
$^{3}$North Carolina State University \\
\texttt{tchen169@asu.edu, pingzhi@cs.unc.edu, zhou22@ncsu.edu,} \\
\texttt{tianlong@cs.unc.edu, hua.wei@asu.edu}
}

\begin{document}
\maketitle

\begin{abstract}
Visual Language Models (VLMs) have gained significant popularity due to their remarkable ability. While various methods exist to enhance privacy in text-based applications, privacy risks associated with visual inputs remain largely overlooked such as Protected Health Information (PHI) in medical images. To tackle this problem, two key tasks: accurately localizing sensitive text and processing it to ensure privacy protection should be performed. To address this issue, we introduce VisShield (Vision Privacy Shield), an end-to-end framework designed to enhance the privacy awareness of VLMs. Our framework consists of two key components: a specialized instruction-tuning dataset OPTIC  (Optical Privacy Text Instruction Collection) and a tailored training methodology. The dataset provides diverse privacy-oriented prompts that guide VLMs to perform targeted Optical Character Recognition (OCR) for precise localization of sensitive text, while the training strategy ensures effective adaptation of VLMs to privacy-preserving tasks. Specifically, our approach ensures that VLMs recognize privacy-sensitive text and output precise bounding boxes for detected entities, allowing for effective masking of sensitive information.  Extensive experiments demonstrate that our framework significantly outperforms existing approaches in handling private information, paving the way for privacy-preserving applications in vision-language models. Our dataset and code can be found here.\footnote{\url{https://github.com/tiejin98/VLM_Deidentification}}.
\end{abstract}    
\section{Introduction}
\label{sec:intro}

Vision Language Models (VLMs)~\cite{alayrac2022flamingo,liu2024visual,bai2023qwen}, which are developed following the impressive success of LLMs, show a remarkable ability to solve image-related tasks. Similar to text-only Large Language Models (LLMs)~\cite{dubey2024llama,abdin2024phi}, which pose potential privacy risks by memorizing and outputting sensitive information from training data~\cite{mireshghallah2022memorization, huang2022large,carlini2021extracting}, VLMs also suffer from privacy risks because VLMs share the generation part with LLMs~\cite{liu2024protecting}.

\begin{figure}[t!]
\centering
\includegraphics[width=0.45\textwidth]{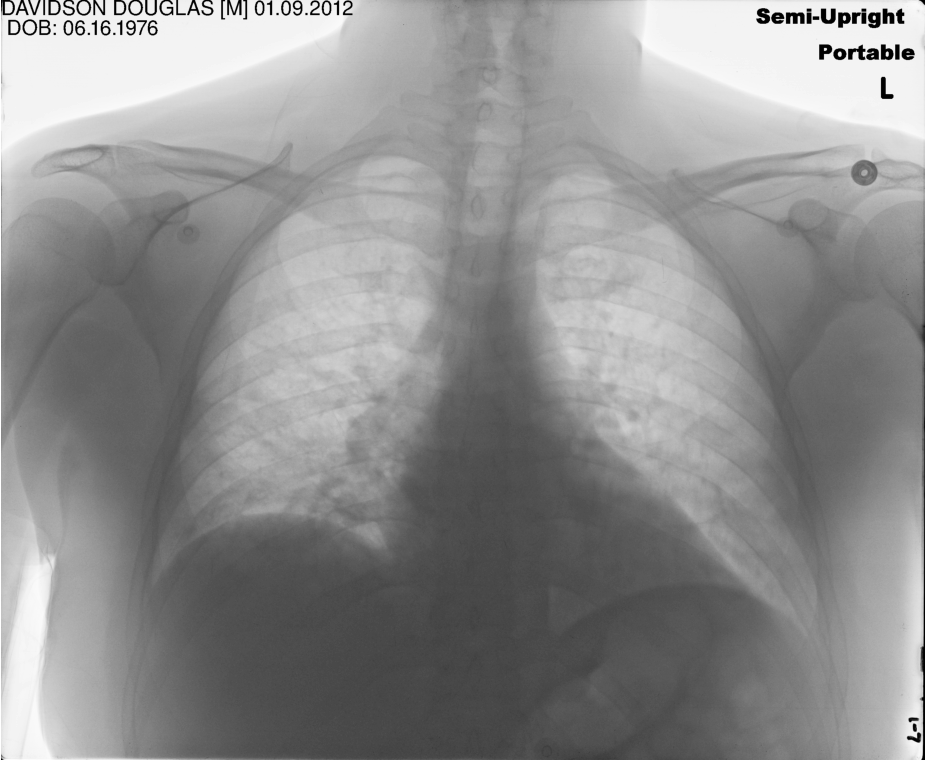}
\caption{An illustrative example of medical imaging containing protected health information~(PHI), shown in the top-left region, adapted from \citet{rutherford2021dicom}. The displayed information is synthetic and thus remains unmasked for demonstration purposes.}
\label{fig:example_privacy}
\vspace{-5mm}
\end{figure}

To mitigate the privacy risks of text-only LLMs, several methods are proposed. For example, \citet{jang2022knowledge} utilized knowledge editing to make LLMs forget the private information. Moreover, \citet{zeng2024privacyrestore} proposed privacy restoration to remove the private information in the input and \citet{yang2024robust} leveraged an auxiliary LLM to remove the sensitive information in the training data. However, most of them focus on the text while neglecting the potentially sensitive information in visual input. For example, medical images often contain protected health information (PHI), which is considered sensitive information. We also show an example of PHI in \cref{fig:example_privacy}. 

\begin{figure*}[t!]
\centering
\includegraphics[width=0.8\textwidth]{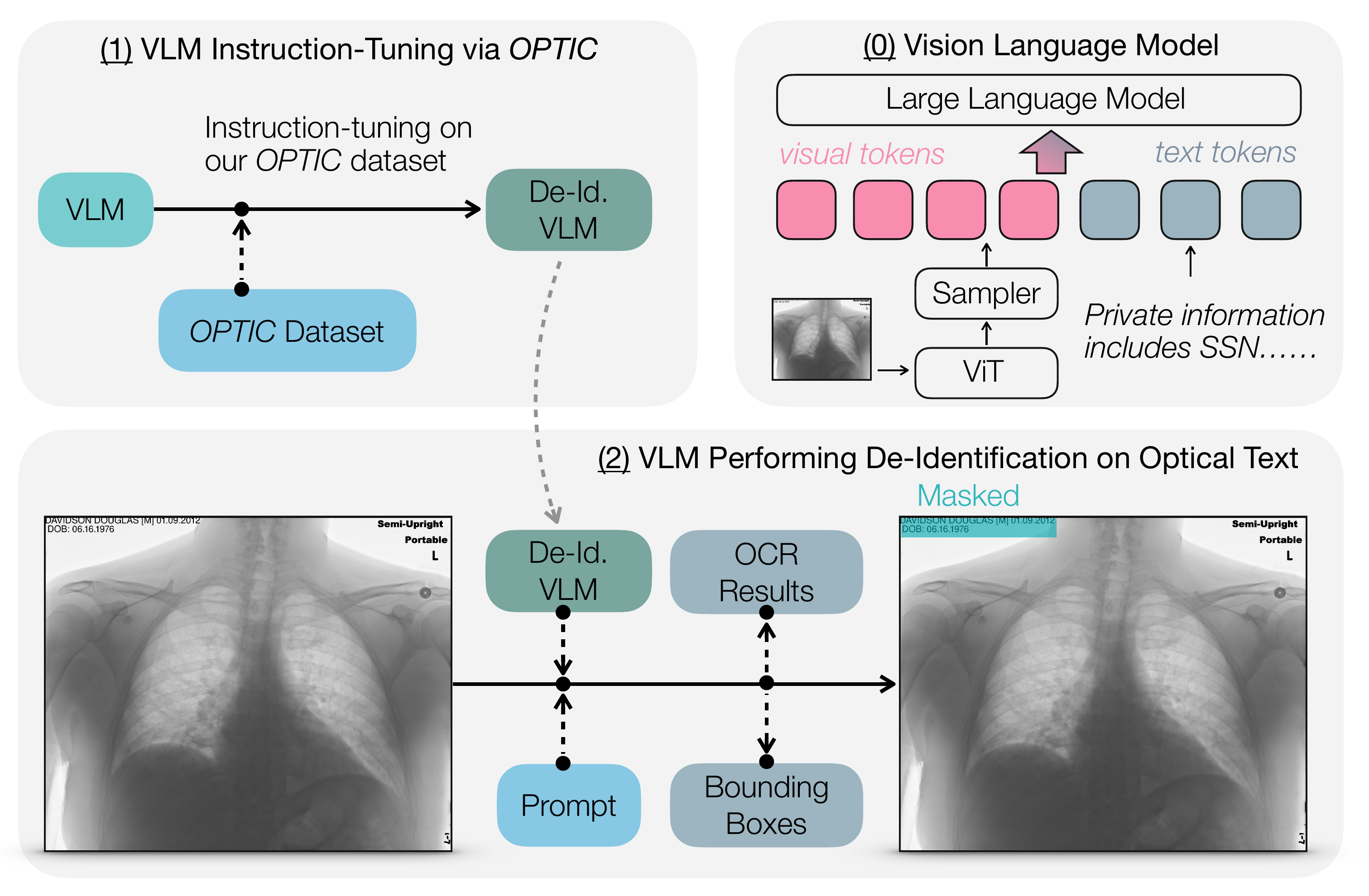}
\vspace{-10pt}
\caption{The proposed de-identification pipeline. Our approach leverages instruction-tuned VLMs to first perform targeted OCR on privacy-sensitive regions, followed by selective masking of identified confidential information.}
\label{fig:de_pipeline}
\vspace{-2mm}
\end{figure*}

To tackle privacy issues arising from vision data, one promising solution is data de-identification~\cite{ribaric2016identification}. De-identification is the process of removing or masking personally identifiable information (PII) from datasets to ensure privacy. However, previous works on image de-identification mainly focus on faces, which aim at obscuring identifiable facial features using generative models~\cite{brkic2017know,cao2021personalized}. There is a lack of work focusing on textual private information in vision data. To the best of our knowledge, only Presidio~\cite{presidio} attempts to de-identify such information. However, Presidio lacks the flexibility to define what constitutes private information and demonstrates suboptimal performance in our experiments.

To address the lack of methods for de-identifying textual private information in vision data, two key tasks are required: accurately localizing sensitive text and processing it to ensure privacy protection. Therefore, in this paper, we propose an end-to-end framework named VisShield~(Vision Privacy Shield), which leverages a Vision Language Model to assist in the de-identification of vision data. Our framework includes two components: \\
\noindent1) A specialized instruction-tuning dataset OPTIC~(Optical Privacy Text Instruction Collection) designed to teach VLMs how to handle privacy-sensitive textual elements. This dataset includes diverse, privacy-oriented instructions that guide VLMs to perform OCR-based localization of private text. We generate synthetic image-text pairs with embedded fake private information, covering both natural and medical image scenarios, ensuring robust generalization. Our dataset comprises 50M samples, providing a rich training resource for localizing sensitive text.\\
\noindent2) A tailored training methodology that enables a VLM to accurately understand customized definitions of private information and apply de-identification mechanisms effectively. We fine-tuned a pre-trained VLM, \texttt{Kosmos-2.5}~\cite{lv2023kosmos} on the OPTIC dataset to enable the VLM to process sensitive text accurately.

Our framework pipeline as shown in \cref{fig:de_pipeline} enables the VLM to understand customized definitions of private information and extract private information through OCR, which can then be masked to ensure privacy. Extensive experiments demonstrate that our VisShield achieves superior privacy-aware OCR performance and leads to potential new applications of VLMs. Overall, we summarize our contribution below:



\begin{itemize}
    \item To the best of our knowledge, we are the first to address the problem of de-identification with customized definitions of textual private information in vision data.
    \item We collect a diverse instruction-tuning dataset, which contains both text and image parts. This dataset comprises up to 50M image-text pairs, enabling VLMs to output OCR results for identifying private information in images.
    \item We fine-tune \texttt{Kosmos-2.5} to demonstrate that even a small portion of our dataset suffices for fine-tuning a pre-trained VLM to assist with de-identification.
\end{itemize}
\section{Related Work}
\label{sec:Related}
\begin{figure}[t!]
\centering
\includegraphics[width=0.49\textwidth]{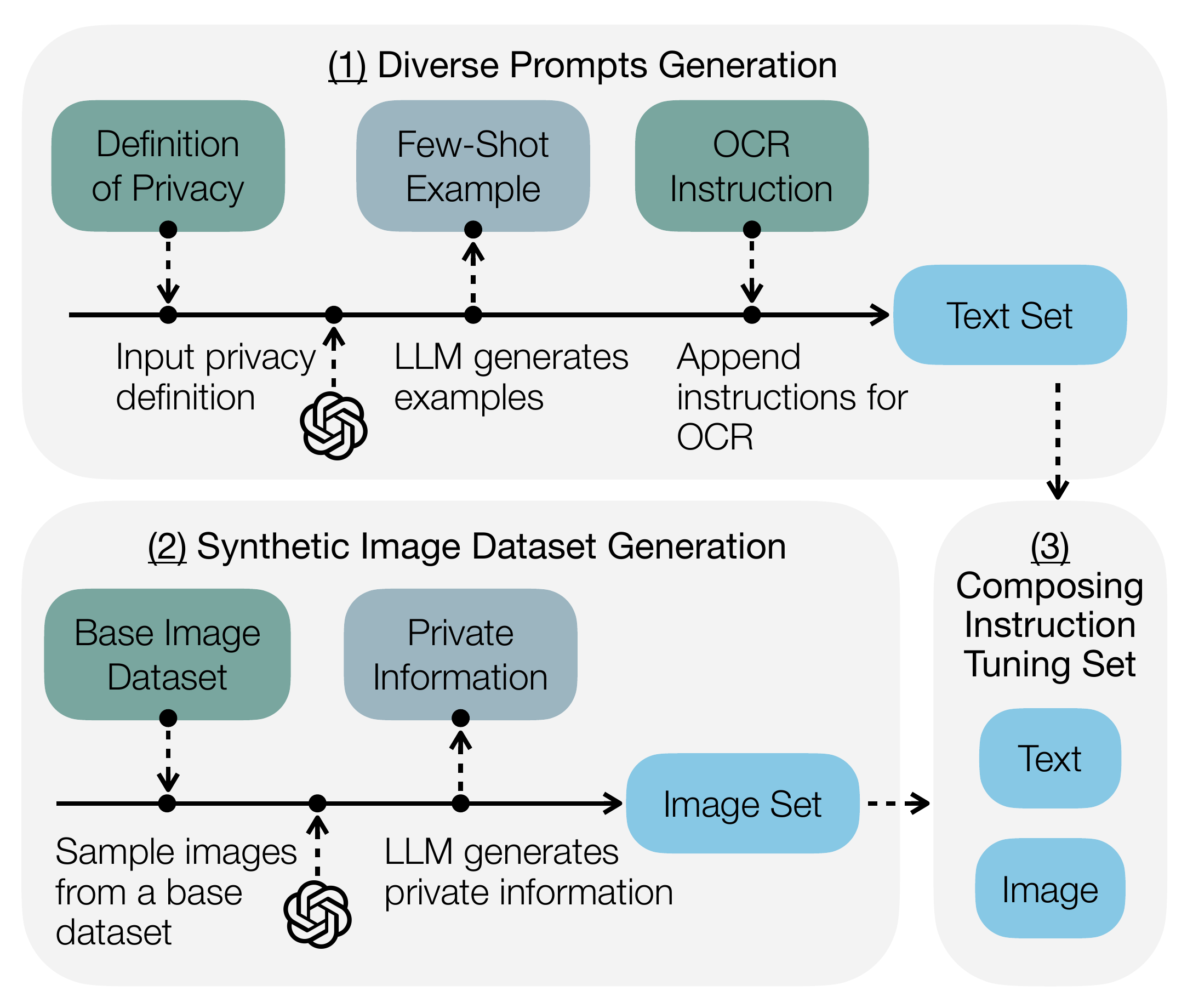}
\caption{Overview of our three-stage dataset generation pipeline: ($1$) leveraging large language models (LLMs) to synthesize diverse instruction prompts, ($2$) creating synthetic images containing private information through controlled generation, and ($3$) producing aligned instruction-label pairs by combining the generated prompts with the synthetic image dataset.}
\label{fig:dateset_pipeline}
\end{figure}

\paragraph{Vinson Language Models}  With the help of LLMs’ powerful reasoning abilities, Vision Language Models (VLMs) have achieved significant success in recent days. Different models, including Llava~\cite{liu2024visual}, BLIP2~\cite{li2023blip}, Flamingo~\cite{alayrac2022flamingo}, Qwen2-VL~\cite{wang2024qwen2}, mini-GPT4~\cite{zhu2023minigpt} have shown their impressive results among different vision-related tasks, which contains but not limited to Visual question answering~\cite{biten2022latr,guo2023images,ozdemir2024enhancing,hu2024bliva}, image captioning~\cite{rotstein2024fusecap,yang2024exploring} or visual grounding~\cite{peng2023kosmos, yu2025merlin}. Among all tasks, document OCR~\cite{wei2025vary,lv2023kosmos} and its application, which outputs the bounding box for texts in the images and answers the question based on the texts, are the task most similar to ours, where our task is based on the bounding boxes for texts. However, none of the previous works have utilized VLMs for de-identification to protect the privacy of vision data. Our collected dataset and model not only address this gap but also expand the application scope of VLMs.

\paragraph{Instruction Tuning} Instruction tuning is used to make language models follow natural language instructions and
complete more complex tasks~\cite{ouyang2022training,wang2022benchmarking,wei2021finetuned,zhang2023instruction}. Instruction tuning improves the zero- and few-shot generalization abilities of LLMs for both text-only LLMs, which include ChatGPT~\cite{achiam2023gpt-4,openai2023chatgpt}, Llama family~\cite{touvron2023llama,dubey2024llama} and Flan family~\cite{longpre2023flan,chung2024scaling}, to VLMs~\cite{liu2024visual,liu2024improved} with diverse vision prompts as additional inputs. 

\begin{figure*}[t!]
\centering
\includegraphics[width=0.92\textwidth]{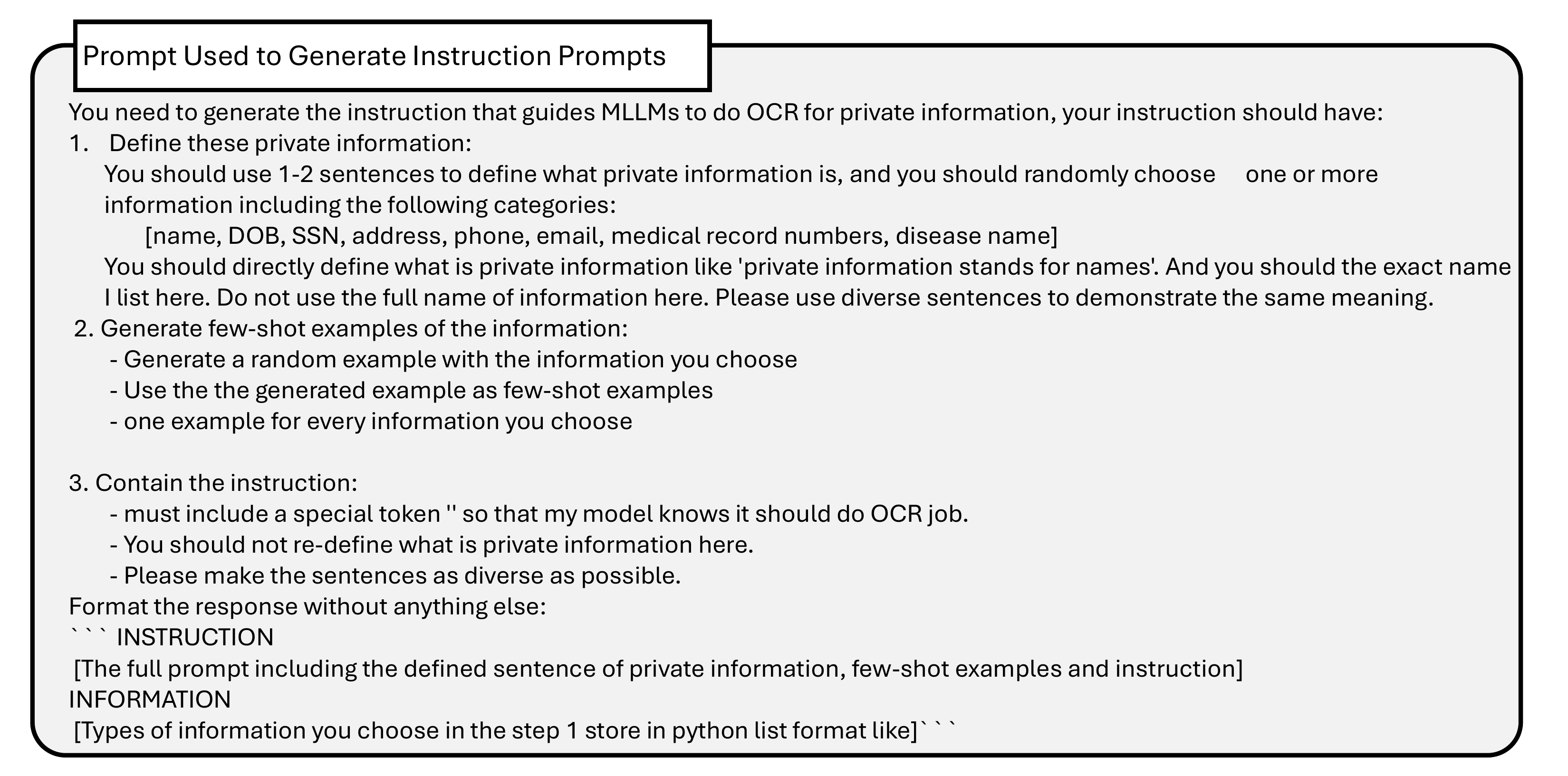}
\caption{Template prompt utilized for instruction generation, implemented with GPT-4 and Claude-3.5 Sonnet. This prompt guides the LLMs to synthesize diverse task-specific instruction prompts.}
\label{fig:GPT_prompt}
\end{figure*}

The quality of instruction tuning is highly dependent on the quality of the tuning dataset~\cite{zhou2024lima}. Therefore, previous works like Llava~\cite{liu2024visual,liu2024improved} leverage LLMs to expand the existing image dataset~\cite{lin2014microsoft} to various instruction-following datasets. In this work, we use a similar pipeline based on the flickr30k dataset~\cite{plummer2015flickr30k} and medical images~\cite{rutherford2021dicom}.


\paragraph{De-identification} De-identification is the process of removing or obfuscating personal information from data to prevent the identification of individuals~\cite{ribaric2016identification}. For image de-identification, most current methods aim at face images, where replacing faces in images to protect privacy~\cite{gross2006model, brkic2017know,cao2021personalized}. However, to the best of our knowledge, there is no previous work focused on de-identifying burn-in pixels (texts in the images), especially with the help of VLMs. Therefore, our model fills the gap and extends the application range of VLMs.



\section{Methodology}
\label{sec:dataset}

\subsection{De-identification Pipeline} 
As shown in \cref{fig:de_pipeline}, our full de-identification pipeline contains prompting fine-tuned VLMs to output OCR results. Then, we mask out the text using the top-left color of every bounding box in the output. To achieve a successful de-identification as shown in the pipeline, two key tasks: 1) accurately localizing sensitive text and 2) processing it to ensure privacy protection are required. To perform these two tasks, we propose a framework called VisShield and introduce two components of VisShield: 1) a specialized dataset OPTIC for instruction tuning and 2) a training methodology.


\subsection{OPTIC Dataset}
Our instruction-tuning approach aims to enable VLMs to analyze and extract private information precisely through OCR. In order to achieve this goal, the OPTIC dataset contains in total of 50M sample sizes with various instruction prompts and images with private information. 

\subsubsection{Instruction Prompts}
\vspace{-3mm}
\begin{table}[htbp]
    \centering
    \resizebox{0.48\textwidth}{!}{
        \begin{tabular}{llp{6cm}}
            \toprule
            \textbf{Config} & \textbf{Numbers} & \textbf{Options} \\
            \midrule
            Font       & 6   & Arial, Times\_New\_Roman, Verdana, courbi, DejaVuSans, NotoSansMono \\
            Font Size  & N/A & 3\%-9\% of the whole image \\
            Font Color & 9   & White, Black, yellow, cyan, orange, pink, \\ 
                       &     & lightgreen, red, blue \\
            \bottomrule
        \end{tabular}
    }
    \caption{Detailed options of different generation configurations. During generation, we will random sample each configuration to ensure a diverse generation.}
    \label{tab:font_config}
    \vspace{-3mm}
\end{table}
The instruction set encompasses four distinct contextual categories, which we detail in the following sections.

\paragraph{Definition of Private Information}
The notion of private information is inherently context-dependent and domain-specific. For instance, numerical sequences in medical contexts may represent confidential medical record identifiers, while similar numerical patterns in other domains might have no privacy implications. We explicitly incorporate contextual definitions within each instruction prompt to enable VLMs to identify and process private information across diverse scenarios accurately. These definitions follow a precise format (e.g., "Private information encompasses names and email addresses") to eliminate ambiguity and ensure consistent interpretation by the model.

\paragraph{Few-shot Examples} Providing abstract definitions of private information alone is often insufficient for optimal VLM performance, as the format and structure of sensitive data vary significantly across contexts. For instance, medical record numbers follow institution-specific formats, while phone number structures differ across national boundaries. To enhance the instruction-following capabilities of VLMs and improve OCR accuracy for targeted information, we leverage in-context learning~\cite{dong2022survey,zhang2023makes} by incorporating carefully curated few-shot examples into our instructions. These examples are specifically designed to align with and contextualize the provided definitions, enabling more robust recognition of diverse data formats.

\paragraph{Instruction} The critical component of our instruction prompts is a targeted directive that guides VLMs to extract OCR results exclusively from private information. We leverage a specialized token \textit{\textless{}ocr\textgreater{}} for OCR tasks. This token is consistently incorporated across all instructions, serving as a standardized trigger that signals the fine-tuned VLM to initiate OCR processing for privacy-relevant content within the prompted region.


\paragraph{Generation} Building upon established methodologies~\cite{liu2024visual,liu2024improved}, we employ state-of-the-art large language models to generate diverse instruction prompts. Specifically, we utilize GPT-4~\cite{openai_gpt4osystemcard} and Claude-3.5 Sonnet~\cite{anthropic_claude35}, which represent the current frontier of language model capabilities. Our framework encompasses eight distinct categories of sensitive information, ranging from personally identifiable information~(PII), such as email addresses and Social Security Numbers~(SSN), to protected health information, including disease classifications. A comprehensive taxonomy of these information types is presented in \cref{tab:information_types}. We developed structured prompts that direct these LLMs to randomly sample from these information categories, generate few-shot examples, and produce diverse task-specific instructions. The complete prompt template used for instruction generation is illustrated in \cref{fig:GPT_prompt}, with a representative example of a generated instruction prompt shown in Appendix \cref{fig:example_prompt}. We have a total of 2500 different instruction prompts, with 1250 generated by GPT-4o and 1250 generated by Claude-3.5-Sonnet.

\begin{table}[h]
    \centering
    \resizebox{0.48\textwidth}{!}{
        \begin{tabular}{llp{6cm}}
            \toprule
            \textbf{Type of Information} & \textbf{Number} & \textbf{Example} \\
            \midrule
            Name            & 16300  & Joe Dohn \\
            DOB             & 16276  & 18 Jun 1983 \\
            SSN             & 16350  & 071-30-5000 \\
            Phone Number    & 16271  & 555-304-8389 \\
            Address         & 16270  & 086 Holt Summit, CT 58671 \\
            Email           & 16149  & 54jmz@hotmail.com \\
            Medical Numbers & 16243  & MRN93987011 \\
            Disease Name    & 16274  & Migraine \\
            \bottomrule
        \end{tabular}
    }
    \caption{Examples of information types we consider in this paper. We consider 8 types with balanced numbers of size in each type. All the information is fake. }
    \label{tab:information_types}
    \vspace{-5mm}
\end{table}

\subsubsection{Synthetic Images}

To fine-tune the VLMs, we need images containing private information and bounding box annotations for the private information in images. However, since we are the first to address the challenge of textual private information in images, there is a lack of existing image datasets. In order to obtain the dataset, we create images with private information based on the base image datasets.

\begin{table*}[h!]
\centering
\resizebox{\textwidth}{!}{
\begin{tabular}{c|cc|cc|cc|cc|cc|cc|cc|cc}
\toprule
\midrule
\multicolumn{1}{c|}{Model} & \multicolumn{2}{c|}{Name} & \multicolumn{2}{c|}{DOB} & \multicolumn{2}{c|}{SSN} & \multicolumn{2}{c|}{Email} & \multicolumn{2}{c|}{Phone Number} & \multicolumn{2}{c|}{Address} & \multicolumn{2}{c|}{Medical Number} & \multicolumn{2}{c}{Disease Name} \\
\midrule & F1 & IoU & F1 & IoU & F1 & IoU & F1 & IoU & F1 & IoU & F1 & IoU & F1 & IoU & F1 & IoU \\
\midrule
\multicolumn{17}{c}{Evaluation Set Generated by Training Base Image Dataset} \\
\midrule
Full & $\mathbf{0.9733}$ & $0.9134$ & $0.9849$ & $0.8984$ & $\mathbf{0.9781}$ & $0.9103$ & $\mathbf{0.9719}$ & $\mathbf{0.9482}$  & 0.9736 & 0.9045 & 0.9809 & 0.9615 & $\mathbf{0.9762}$ & 0.8626 & 0.9426 & $\mathbf{0.8920}$ \\
LoRA & $0.9728$ & $\mathbf{0.9194}$ & $0.9849$ & $\mathbf{0.9196}$ & $0.9714$ & $\mathbf{0.9205}$ & $0.9601$ & $0.9419$ & $\mathbf{0.9801}$ & $\mathbf{0.9144}$ & $\mathbf{0.9849}$ & $\mathbf{0.9690}$ & $0.9714$ & $\mathbf{0.8898}$ & $\mathbf{0.9501}$ & $0.8782$ \\
Presidio & N/A & 0.0085 & N/A & 0.0074 & N/A & 0.0067 & N/A & 0.0119 & N/A & 0.0072 & N/A & 0.0141 & N/A & 0.0074 & N/A & 0.0067  \\
\midrule
\multicolumn{17}{c}{Evaluation Set Generated by COCO} \\
\midrule
Full & 0.9708 & 0.9058 & $\mathbf{0.9903}$ & $\mathbf{0.9472}$  & 0.9767 & 0.8997 & $\mathbf{0.9693}$ & 0.9338 & $\mathbf{0.9838}$ & 0.9017 & 0.9703 & 0.9632 & 0.9637 & 0.8706 & 0.9565 & 0.8805 \\
LoRA & $\mathbf{0.9713}$ & $\mathbf{0.9075}$ & $0.9818$ & $0.9083$ & $\mathbf{0.9859}$ & $\mathbf{0.9157}$ & $0.9679$ & $\mathbf{0.9369}$ & $0.9772$ & $\mathbf{0.9097}$ & $\mathbf{0.9802}$ & $\mathbf{0.9657}$ & $\mathbf{0.9818}$ & $\mathbf{0.8995}$ & $\mathbf{0.9661}$ & $\mathbf{0.8764}$ \\
Presidio & N/A & 0.0067 & N/A & 0.0060 & N/A  & 0.0054 & N/A  & 0.0085 &  N/A & 0.0057 & N/A & 0.1201 & N/A & 0.0057 & N/A & 0.0052 \\
\midrule
\multicolumn{17}{c}{Evaluation Set Generated by ADE-20K} \\
\midrule
Full & $\mathbf{0.9499}$ & $\mathbf{0.9075}$ & $\mathbf{0.9842}$ & $\mathbf{0.8849}$ & 0.9576 & $\mathbf{0.8918}$ & $\mathbf{0.9718}$ & 0.9252 & $\mathbf{0.9481}$  & $\mathbf{0.9200}$ & $\mathbf{0.9564}$ & $\mathbf{0.9508}$ & $\mathbf{0.9818}$ & 0.8633 & 0.9606 & 0.8863 \\
LoRA & $0.9300$ & $0.8921$ & $0.9769$ & $0.9025$ & $\mathbf{0.9740}$ & $0.8913$ & $0.9496$ & $\mathbf{0.9282}$ & $0.9412$ & $0.8984$ & $0.9513$ & $0.9453$ & $0.9725$ & $\mathbf{0.8655}$ & $\mathbf{1.0000}$ & $\mathbf{0.8905}$ \\
Presidio & N/A  & 0.0027 & N/A & 0.0024 & N/A & 0.0021 & N/A & 0.0033 & N/A & 0.0022 & N/A & 0.0048 & N/A & 0.0023 & N/A & 0.0021 \\
\midrule
\multicolumn{17}{c}{Evaluation Set Generated by RITE} \\
\midrule
Full & 0.9836 & 0.9251 & 0.9633 & 0.9093 & $\mathbf{0.9863}$ & 0.9149 & $\mathbf{0.9842}$ & 0.9449 & $\mathbf{0.9911}$ & 0.9176 & $\mathbf{0.9910}$ & $\mathbf{0.9751}$ & $\mathbf{0.9902}$ & 0.8777 & $\mathbf{1.0000}$ & 0.9058 \\
LoRA & $\mathbf{0.9938}$ & $\mathbf{0.9723}$ & $\mathbf{0.9851}$ & $\mathbf{0.9785}$ & $0.9843$ & $\mathbf{0.9953}$ & $0.9689$ & $\mathbf{0.9669}$ & $0.9109$ & $\mathbf{0.9304}$ & $0.9266$ & $0.9491$ & $0.9210$ & $\mathbf{0.9760}$  & $0.8966$ & $\mathbf{0.9118}$ \\
Presidio & N/A & 0.0077 & N/A & 0.0070 & N/A & 0.0066 & N/A & 0.0096 & N/A & 0.0073 & N/A & 0.0126 & N/A & 0.0068 & N/A & 0.0062 \\
\midrule
\bottomrule
\end{tabular}
}
\caption{Comparative analysis of model performance across information categories, model architectures, and evaluation datasets. We evaluate using randomly sampled instruction prompts from the training set. Results demonstrate that our fine-tuned models achieve strong generalization capabilities, with full model fine-tuning consistently outperforming other adaptation strategies.}
\label{tab:different_image}
\vspace{-3mm}
\end{table*}

\vspace{-2mm}
\paragraph{Base Image Dataset} We overlay private information onto the base image dataset to generate vision data, where the base image dataset plays an important role. We hope the base image dataset includes diverse images to enhance generalization ability. Therefore,  we first utilize the existing dataset that already has diverse images from image caption domains. In detail, we use the flickr30k dataset~\cite{plummer2015flickr30k} as the first part of the base image dataset. Additionally, we include the medical images in our base image dataset since the medical area is the most important application area for de-identification. Specifically, we use a public medical dataset containing various types of medical images from \citet{rutherford2021dicom}.


\vspace{-2mm}
\paragraph{Generation} For the generation of our synthetic dataset, we first sample one base image from our base image datasets and then overlay the private information on the sampled image. In detail, after sampling the image, we determine the amount of private information to be overlaid on the sampled image by randomly selecting an integer between four and ten. Then for each piece of information, we randomly decide the type of the information and generate fake information using the Faker package~\cite{faker}. Then, we print the generated fake information on the sampled image using PIL package~\cite{Pillow}, which also provides the ground truth bounding box information for the text. While overlaying the information on the sampled image, we use different fonts, font sizes, and colors to ensure the diversity of generated text. The details of the generation configuration can be found at \cref{tab:font_config}. In total, we generate 20,000 images with more than 130,000 bounding boxes.

\subsubsection{Label Generation} So far, we have introduced the input part of our dataset. However, to fine-tune VLMs, we also need labels to optimize the loss function. Our target is to make VLMs output the OCR results for the defined private information. The labels should differ based on the same instruction prompt with different images or for different instruction prompts applied to the same image. Therefore, we first randomly sample one prompt from instruction prompts and one image from the synthetic image dataset to form the full input and then generate the label corresponding to the full input. We provide bounding boxes only for the private information types that are used to define private information in the instruction to generate labels. For example, if the instruction prompt specifies that 'private information only stand for names', then we will only provide bounding box for names in the given image as the label. If there is no such information in the image, the answer will be 'No private information'. If there is such information, the answer will be the concatenation of each bounding box which is expressed as \textit{\textless bbox\textgreater{} $\textless x_{tl} \textgreater$ $\textless y_{tl} \textgreater$ $\textless x_{br} \textgreater$ $\textless y_{br} \textgreater$ \textless /bbox\textgreater{}}. The coordinates denote the top-left and bottom-right corners of the bounding box. 
 
\subsection{Training on OPTIC}

While the OPTIC dataset provides a rich foundation for training privacy-aware VLMs, effectively leveraging it to improve the model's capability remains a significant challenge. To address this challenge, we introduce our training strategy and our strategy is built upon three key principles:
\vspace{-2mm}
\paragraph{Efficiency}  
While our dataset contains 50M samples, training on the full dataset is computationally expensive and unnecessary. Instead, we demonstrate that training on a \textbf{small subset of 100K samples} is sufficient to significantly enhance the model's de-identification capabilities. This approach allows us to reduce resource requirements.
\vspace{-2mm}

\paragraph{Knowledge Transfer} 
Instead of training a VLM from scratch, we fine-tune \texttt{Kosmos-2.5}~\cite{lv2023kosmos}, a pre-trained multimodal model that inherently supports OCR extraction from images. However, to make it privacy-aware, our fine-tuning process could improve its ability to selectively extract only privacy-relevant text rather than all OCR content, and refine its bounding box localization for privacy-sensitive elements.
\vspace{-2mm}

\paragraph{Adaptation Strategies}
We explore two fine-tuning strategies to integrate privacy-awareness into the model. The first is \textbf{full fine-tuning}, where the entire model is fine-tuned on privacy-sensitive OCR tasks, while the second is \textbf{LoRA}~\cite{hu2021lora}, a parameter-efficient approach that updates only a limited set of trainable parameters, reducing memory consumption.

With our training strategy, we ensure that our end-to-end framework learns to effectively identify, localize, and process private textual information.



\section{Experiments}
\label{sec:experiments}

\begin{table*}[h!]
\centering
\resizebox{\textwidth}{!}{
\begin{tabular}{c|cc|cc|cc|cc|cc|cc|cc|cc}
\toprule
\midrule
\multicolumn{1}{c|}{Model} & \multicolumn{2}{c|}{Name} & \multicolumn{2}{c|}{DOB} & \multicolumn{2}{c|}{SSN} & \multicolumn{2}{c|}{Email} & \multicolumn{2}{c|}{Phone Number} & \multicolumn{2}{c|}{Address} & \multicolumn{2}{c|}{Medical Number} & \multicolumn{2}{c}{Disease Name} \\
\midrule
& F1 & IoU & F1 & IoU & F1 & IoU & F1 & IoU & F1 & IoU & F1 & IoU & F1 & IoU & F1 & IoU \\
\midrule
\multicolumn{17}{c}{Instruction Prompts Generated by Gemma1.5} \\
\midrule
Full & 0.9493 & 0.9008 & 0.9636 & 0.9013 & 0.9842 & 0.9075 & 0.9537 & 0.9290 & 0.9114 & 0.9080 & 0.9591 & 0.9644 & 0.9760 & 0.8586 & 0.9247 & 0.8973 \\
LoRA & $0.9561$ & $0.9791$ & $0.9764$ & $0.9491$ & $0.9721$ & $0.9798$ & $0.9669$ & $0.9767$ & $0.8960$ & $0.9121$ & $0.9177$ & $0.9429$ & $0.9130$ & $0.9721$ & $0.8815$ & $0.8948$ \\
Presidio & $N/A$ & 0.0085 & $N/A$ & 0.0074 & $N/A$ & 0.0067 & $N/A$ & 0.0119 & $N/A$ & 0.0072 & $N/A$ & 0.0141 & $N/A$ & 0.0074 & $N/A$ & 0.0067  \\
\midrule
\multicolumn{17}{c}{Instruction Prompts Generated by Human} \\
\midrule
Full & $0.9420$ & $0.9247$ & $0.9943$ & $0.9094$ & $0.9723$ & $0.9211$ & $0.9129$ & $0.9353$ & $0.9842$ & $0.9010$ & $0.9823$ & $0.9613$ & $0.9511$ & $0.8749$ & $0.9746$ & $0.9210$ \\
LoRA & $0.9758$ & $0.9667$ & $0.9847$ & $0.9499$ & $0.9799$ & $0.9560$ & $0.9414$ & $0.9877$ & $0.9196$ & $0.9251$ & $0.9247$ & $0.9447$ & $0.9333$ & $0.9675$ & $0.8751$ & $0.8911$ \\
Presidio & $N/A$ & 0.0085 & $N/A$ & 0.0074 & $N/A$ & 0.0067 & $N/A$ & 0.0119 & $N/A$ & 0.0072 & $N/A$ & 0.0141 & $N/A$ & 0.0074 & $N/A$ & 0.0067  \\
\midrule
\bottomrule
\end{tabular}
}
\caption{Performance comparisons for different types of information, different models, and different instruction prompts. The evaluation image set is chosen for the evaluation set generated by the training base image dataset.}
\label{tab:different_prompts_base_image}
\end{table*}

\begin{table*}[h!]
\centering
\resizebox{\textwidth}{!}{
\begin{tabular}{c|cc|cc|cc|cc|cc|cc|cc|cc}
\toprule
\midrule
\multicolumn{1}{c|}{Model} & \multicolumn{2}{c|}{Name} & \multicolumn{2}{c|}{DOB} & \multicolumn{2}{c|}{SSN} & \multicolumn{2}{c|}{Email} & \multicolumn{2}{c|}{Phone Number} & \multicolumn{2}{c|}{Address} & \multicolumn{2}{c|}{Medical Number} & \multicolumn{2}{c}{Disease Name} \\
\midrule
& F1 & IoU & F1 & IoU & F1 & IoU & F1 & IoU & F1 & IoU & F1 & IoU & F1 & IoU & F1 & IoU \\
\midrule
\multicolumn{17}{c}{Instruction Prompts Generated by Gemma1.5} \\
\midrule
Full & 0.9483 & 0.9062 & 0.9625 & 0.8985 &0.9771 & 0.9000 & 0.9309 & 0.8990 & 0.9245 &0.9090  & 0.9782 & 0.9625 & 0.9464 & 0.8673 & 0.8586 & 0.8942 \\
LoRA & $0.9852$ & $0.9689$ & $0.9851$ & $0.9636$ & $0.9576$ & $0.9751$ & $0.9635$ & $0.9749$ & $0.9017$ & $0.9078$ & $0.9105$ & $0.9309$ & $0.9100$ & $0.9669$ & $0.8915$ & $0.8906$ \\
Presidio & $N/A$ & 0.0067 & $N/A$ & 0.0060 & $N/A$  & 0.0054 & $N/A$  & 0.0085 &  $N/A$ & 0.0057 & $N/A$ & 0.1201 & $N/A$ & 0.0057 & $N/A$ & 0.0052  \\
\midrule
\multicolumn{17}{c}{Instruction Prompts Generated by Human} \\
\midrule
Full & 0.9586 &0.9027 & 0.9928 & 0.9042 &  0.9636 & 0.9153 & 0.9234 & 0.9389 & 0.9697 & 0.9132 & 0.9129 & 0.9626 & 0.9391 & 0.8786 & 0.9139 & 0.8902 \\
LoRA  & $0.9761$ & $0.9826$ & $0.9879$ & $0.9621$ & $0.9602$ & $0.9564$ & $0.9695$ & $0.9727$ & $0.9026$
& $0.9094$ & $0.9139$ & $0.9337$ & $0.9225$ & $0.9668$ & $0.8980$ & $0.9004$ \\
Presidio & $N/A$ & 0.0067 & $N/A$ & 0.0060 & $N/A$  & 0.0054 & $N/A$  & 0.0085 &  $N/A$ & 0.0057 & $N/A$ & 0.1201 & $N/A$ & 0.0057 & $N/A$ & 0.0052 \\
\midrule
\bottomrule
\end{tabular}
}

\caption{Performance comparisons for different types of information, different models, and different instruction prompts. The evaluation image set is chosen for the evaluation set generated by COCO.}
\label{tab:different_prompts_CoCo}
\vspace{-5mm}
\end{table*}

In this section, we provide our experimental results to show the robustness of fine-tuned models. We start with the experimental setting at first.

\subsection{Experimental Setting}

\noindent\textbf{Dataset} To evaluate the robustness and generalization ability of the fine-tuned model, we test the fine-tuned models with five different datasets: 1) Images generated from the same base image dataset and the same instruction prompts in the training set, 2) Images from the same base image dataset and different instruction prompts from the training set, 3) Images from different base image dataset and different instruction prompts from the training set, 4) Images from different base image dataset with extra private information (not in 8 types of private information considered in training) and different instruction prompts from the training set, and 5) real-world images, which is annotated by human as described in \cite{orekondy2018connecting}. We will provide a more detailed introduction to these datasets in the following section.


\noindent\textbf{Training Parameters} For full fine-tuning, we use an epoch of 5, learning rate 2e-5 with batch size 16. For LoRA, following previous work~\cite{sun2023comparative}, we use a larger learning rate 3e-4 and a larger epoch 10 with the same batch size. For both trainings, we use AdamW~\cite{loshchilov2017decoupled} as the optimizer. All training methods are conducted on a single Nvidia Tesla A100 80GB GPU.

\noindent\textbf{Metrics} In this paper, we mainly consider two different metrics to measure the quality. Following previous works~\cite{olejniczak2022text,ren2016faster}, we use F1 to evaluate the quality of OCR results for defined private information and use the Intersection over Union~(IoU) to evaluate the quality of detection, which are both important for the following mask out procedure.

\noindent\textbf{Research Questions} In this section, we mainly focus on three different research questions about the generalization ability of the fine-tuned Model: 1) Whether fine-tuned VLM is stable for different images, 2) Whether fine-tuned VLM is stable for various instructions and 3)Whether the fine-tuned VLM is stable for new information types. Besides, Our experimental results also show that our fine-tuned VLM performs well even in real-world data and we put the detailed results in Appendix.

\subsection{RQ1: Whether Fine-tuned VLM is Stable for Different Images}
To answer this research question, we use different base image datasets to generate the evaluation set. We only provide the results for our method in most cases. In detail, we consider using: 1) our training base image dataset, 2) COCO~\cite{lin2014microsoft}, 3) ADE20K~\cite{zhou2017scene}, and 4) RITE~\cite{hu2013automated} to generate evaluation image datasets, ensuring comprehensive scenarios from city scenes to medical images considered in the experiments. We generate 1500 images for each dataset with the same generation methods but more generation configurations. We compare our model with Presidio~\cite{presidio}, which first uses an OCR engine to extract all possible lines of text from an image. It then applies a local recognizer. The results are shown in \cref{tab:different_image}. The F1 score for Presidio is $N/A$ because it cannot output OCR results. We have the following observations:

\noindent1) The previous tool Presidio shows a bad performance. Since we cannot customize the private definition for Presidio, the performance of Presidio is highly random for different types of information.\\
\noindent2) Our fine-tuned model shows a very good performance with a mean IoU larger than 0.9. And this good performance remains for various image datasets, showing the robustness of our method.\\
\noindent3) There is no clear winner for full fine-tuning and LoRA. Though the LoRA model wins more times, this winning is marginal given the good performance of both models.

\subsection{RQ2: Whether Fine-tuned VLM is Stable for Various Instructions}
To answer the research question related to various instructions, we generate instruction prompts that are different from our training set by involving human writers and Gemini~\cite{team2023gemini}, and then pair the new prompts with three image datasets we used before with one-shot examples. We generate 1500 text-image pairs for model evaluation, and the results are shown in \cref{tab:different_prompts_base_image} and \cref{tab:different_prompts_CoCo}. We have the following observations:\\
\noindent1) Compared with the results in \cref{tab:different_image}, the performance of both full fine-tuning and LoRA exhibits a slight decrease. However, this decrease is minimal, and the fine-tuned models continue to deliver strong performance. \\
\noindent2) Even when using a different image dataset and Instruction Prompts together, our models still achieve strong performance for the de-identification task.

\subsection{RQ3: Whether Fine-tuned VLM is Stable for New Information Type.}
\label{sec:RQ3}
Now, we conduct experiments to test the performance of fine-tuned VLM on new information types. Here, we focus on two new types of information: 1) phone numbers with a format of 11 digits and 2) passport number that begins with a letter and ends with eight numbers. We use a similar method to generate the evaluation set and we regenerate the instruction prompts with the one-shot prompt to ask models to output OCR results for new types of information. We present our results in \cref{tab:information_types}. We find that: \\
\noindent1) Overall, our fine-tuned models continue to demonstrate strong performance when incorporating new types of information, further highlighting their robustness and reliability.\\
\noindent2) Compared to 11-digit phone numbers, the performance on passport numbers is lower because our models had not previously encountered the format of passport numbers. In contrast, earlier phone numbers share a similar pattern with the new ones, aiding the model's performance.

\begin{table}[h!]
\centering
\resizebox{0.45\textwidth}{!}{
\begin{tabular}{c|cc|cc}
\toprule
\midrule
\multicolumn{1}{c|}{Model} & \multicolumn{2}{c|}{11-Digit Phone Number} & \multicolumn{2}{c}{Passport Number} \\
\midrule
& F1 & IoU & F1 & IoU \\
\midrule
\multicolumn{5}{c}{Evaluation Set Generated by Training Base Image Dataset} \\
\midrule
Full & $0.9803$ & $0.8724$ & $0.8887$ & $0.8596$ \\
LoRA & $0.9803$ & $0.8887$ & $0.8725$ & $0.8597$ \\
Presidio & $N/A$ & 0.0071 & $N/A$  & 0.0064 \\
\midrule
\multicolumn{5}{c}{Evaluation Set Generated by COCO} \\
\midrule
Full & 0.9796 & 0.8679 & 0.8920 & 0.8625 \\
LoRA & $0.9023$ & $0.8167$ & $0.8776$ & $0.8583$ \\
Presidio & $N/A$ & 0.0086 & $N/A$ & 0.0054  \\
\midrule
\multicolumn{5}{c}{Evaluation Set Generated by RITE} \\
\midrule
Full & 0.9910 & 0.8761 & 0.9271 & 0.8758 \\
LoRA & $0.8678$ & $0.7463$ & $0.8892$ & $0.8700$ \\
Presidio & $N/A$ & 0.0075 & $N/A$ & 0.0069 \\
\midrule
\bottomrule
\end{tabular}
}
\caption{Performance comparisons for new types of information, different models, and different evaluation image sets. }
\label{tab:new_type}
\vspace{-5mm}
\end{table}



\subsection{Ablation Study}

\begin{figure}[t!]
\centering
\includegraphics[width=0.4\textwidth]{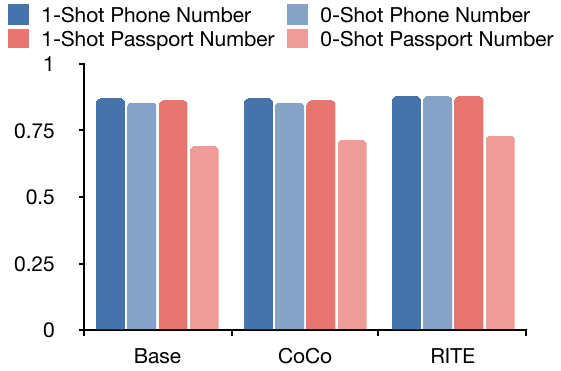}
\caption{IoU performance comparison with different Dataset on 11-digit Phone Number and Passport Number. The experiments are on the full fine-tuned model.}
\label{fig:zero_shot}
\vspace{-5mm}
\end{figure}

In this section, we provide a comparison of the performance of one-shot prompts and zero-shot prompts. More ablation study results can be found in the Appendix. Here, we consider the 11-digit Phone Number and Passport Number as in \cref{sec:RQ3}, and the results for various datasets are presented in \cref{fig:zero_shot}. We found that:

\noindent1) Compared with the one-shot prompt, using the zero-shot prompt can lead to better performance across different datasets, highlighting the importance of few-shot examples.\\
\noindent2) The performance gap between the two prompts is larger when we consider passport numbers. This is because the model has seen similar phone numbers during training, but it has never encountered anything similar to passport numbers before. This highlights the importance of few-shot examples.

\vspace{-1mm}

\subsection{Comparison with OCR}
In previous experiments, we have shown the effectiveness of fine-tuned VLM. However, to solve the problem, there is another training-free method, which first uses an OCR to extract the text and uses another language model to analyze whether we should mask it or not. To test the performance of this kind of method, we compare our fine-tuned model with Tesseract~\cite{smith2007overview} as OCR and Llama2-7B~\cite{touvron2023llama} as the language model. We present the results in \cref{tab:ocr_llm_comparison} on the name category with the test set generated by Base Image Dataset. From the results, we could see that our end-to-end method offers a much better performance. Besides, using OCR with LLM cannot deal with challenging scenarios such as detecting private information in a paragraph. 

\begin{table}[h!]
\small
\centering
\begin{tabular}{lcc}
\toprule
\textbf{Method} & \textbf{F1} & \textbf{IoU} \\
\midrule
Ours-Full & 0.9733 & 0.9134 \\
Tesseract + Llama2-7B & 0.6961 & 0.6728 \\
\bottomrule
\end{tabular}
\caption{Comparison of OCR plus LLM Methods with our method on F1 and IoU Metrics on the name category and  Base Image Dataset.}
\label{tab:ocr_llm_comparison}
\vspace{-3mm}
\end{table}

\subsection{Performance on challenging scenarios}

In the real world, de-identify the private information could be even harder due to the different reasons such as hand-written data or private information in the sentence. As mentioned in the previous section, simply using OCR and LLMs can hardly deal with it. Therefore, to test the further generalization of the proposed method. We mainly conduct the following two experiments:

\noindent \textbf{Experiments on hand-written texts.} To test if the fine-tuned model could recognize the hand-written texts, we form a small-scale evaluation set with 20 images from COCO. Each image contains hand-written text with phone number,email and SSN. The results of full fine-tuned model on these images can be found at \cref{tab:hand_written}. From the result, we could see that the fine-tuned model could still perform very well, which shows the effectiveness.

\begin{figure}[t!]
\centering
\includegraphics[width=0.49\textwidth]{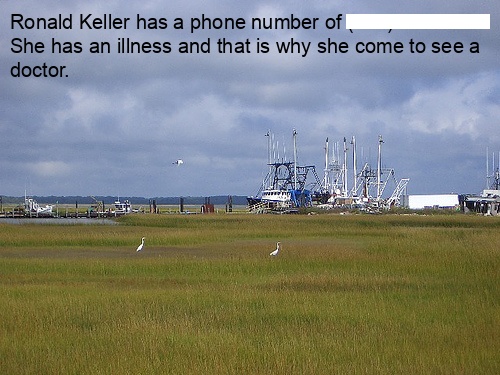}
\caption{An example of de-identification of private information in the sentence. This successful example shows that flexibility of our method.}
\label{fig:example_sentence}
\vspace{-3mm}
\end{figure}

\noindent \textbf{Experiments on sentence.} To test if the fine-tuned model could recognize private information inside the sentence without affecting the other part. In \cref{fig:example_sentence}, we show a case to demonstrate how fine-tuned model performs such scenario, where the private information is defined as phone numbers. From the case, we could see that fine-tuned model successfully recognize the private information in the sentence.

\label{sec:Related}

\begin{table}[ht!]
\small
\centering
\begin{tabular}{lcc}
\toprule
\textbf{Category} & \textbf{F1} & \textbf{IoU} \\
\midrule
Phone Number & 0.9439 & 0.9042 \\
SSN & 0.9377 & 0.9101 \\
Email & 0.8913 & 0.8769 \\
\bottomrule
\end{tabular}
\caption{Performance on full fine-tuned model for images with hand-written text.}
\label{tab:hand_written}
\vspace{-3mm}
\end{table}

\section{Conclusion}
In conclusion, this work presents a novel approach to de-identify textual information in visual data by leveraging the power of VLMs. We generate a comprehensive instruction-tuning dataset with diverse images and instruction prompts. By fine-tuning \texttt{Kosmos-2.5} with this comprehensive instruction-tuning dataset, we demonstrated that VLMs can effectively identify and mask private information. Our results show strong generalization and robustness across different datasets and real-world scenarios, laying a foundation for safer integration of VLMs into privacy-sensitive applications.


\section*{Limitation}
While our approach demonstrates strong performance, it has two key limitations. First, the model's effectiveness depends on the quality of the instruction-tuning dataset, and while we have ensured diversity, rare or highly domain-specific private information formats may still pose challenges. Second, our method relies on OCR accuracy for text extraction, meaning that errors in detecting or recognizing text in low-quality or distorted images could affect de-identification performance. 

\section*{Acknowledgment}
The work was partially supported by NSF awards \#2421839, NAIRR \#240120, \#CNS2431516. This work used AWS through Amazon Research Awards and the CloudBank project supported by National Science Foundation grant \#1925001. Pingzhi Li and Tianlong Chen are partially supported by Amazon Research Award, Cisco Faculty Award, UNC Accelerating AI Awards, NAIRR Pilot Award, OpenAI Researcher Access Award, and Gemma Academic Program GCP Credit Award. The views and conclusions contained in this paper are those of the authors and should not be interpreted as representing any funding agencies.

\bibliography{custom}

\clearpage
\setcounter{page}{1}
\appendix

\section{Example of Instruction prompt}
\begin{figure*}[t!]
\centering
\includegraphics[width=0.8\textwidth]{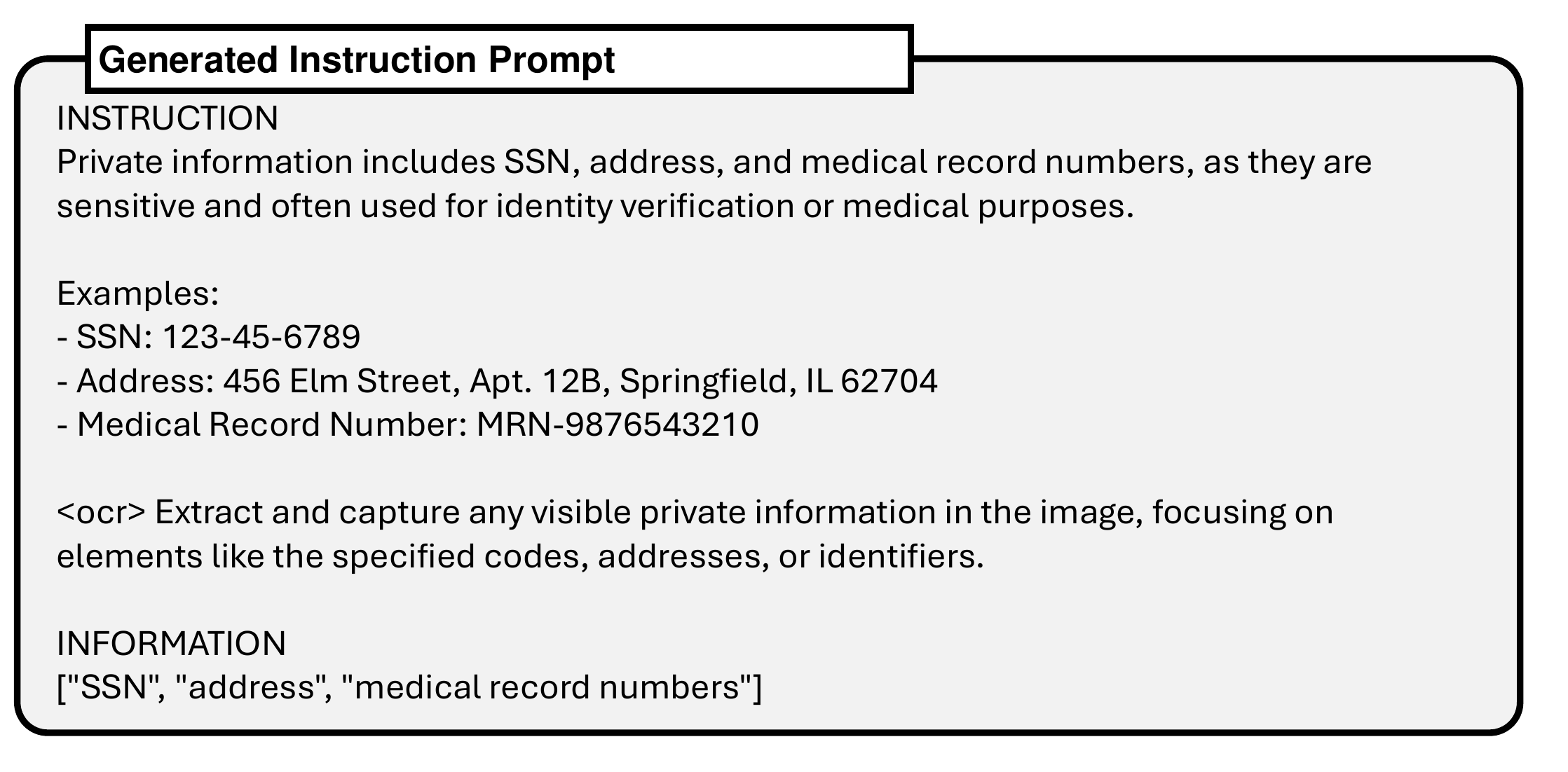}
\caption{One instruction prompt example generated by GPT-4o.}
\label{fig:example_prompt}
\end{figure*}

\section{More Experiments}
\label{sec:experi_supp}

In this section, we provide more experimental results to support our conclusion.

\subsection{mAP Results}

Here, we provide the results for mean Average Precision (mAP) to further demonstrate the results of our experiments. Following previous works in detection, we consider a correction if $\text{IoU} > 0.5$. And the results for different images are provided in \cref{tab:different_image_map} and \cref{tab:different_prompts_base_image_map}. The results in both experiments show that our fine-tuned models also have a very good mAP result, which is reasonable since our IoU results are very high. 

\subsection{Experiments on Real-world Data}
In this section, we use real-world data to test the robustness of the fine-tuned models. In detail, we use images from \cite{orekondy2018connecting}, which contains real-world images from different scenarios. And human annotators will annotate the images with private information and the corresponding bounding box information. More specifically, we focus on names and phone numbers. Then, we use instructions that define private information as names and phone numbers to test the performance on real-world data. Our results can be found in \cref{tab:real_world}. Our experimental results show that even though the performance drops, our full fine-tuned model can also perform well in real-world data, showing good robustness of the model fine-tuned with our dataset.

\begin{table}[h!]
\centering
\resizebox{0.45\textwidth}{!}{
\begin{tabular}{p{1.5cm}|p{1.3cm}p{1.3cm}|p{1.3cm}p{1.3cm}}
\toprule
\multicolumn{1}{c|}{Model} & \multicolumn{2}{c|}{Phone Number} & \multicolumn{2}{c}{Name} \\
\midrule
& F1 & mAP & F1 & mAP \\
\midrule
Full & 0.7001  & 0.5439 & 0.7229 &  0.6037 \\
Presidio & N/A & 0.0002 & N/A & 0.0003 \\
\bottomrule
\end{tabular}
}
\caption{Performance comparisons for different types of information, different models on a real-world dataset}
\label{tab:real_world}
\end{table}

\subsection{More ablation studies}

In this section, we provide more results of our ablation studies. In detail, we provide the results for different numbers of few-shot examples and different training sizes. 

\begin{figure}[t!]
\centering
\includegraphics[width=0.48\textwidth]{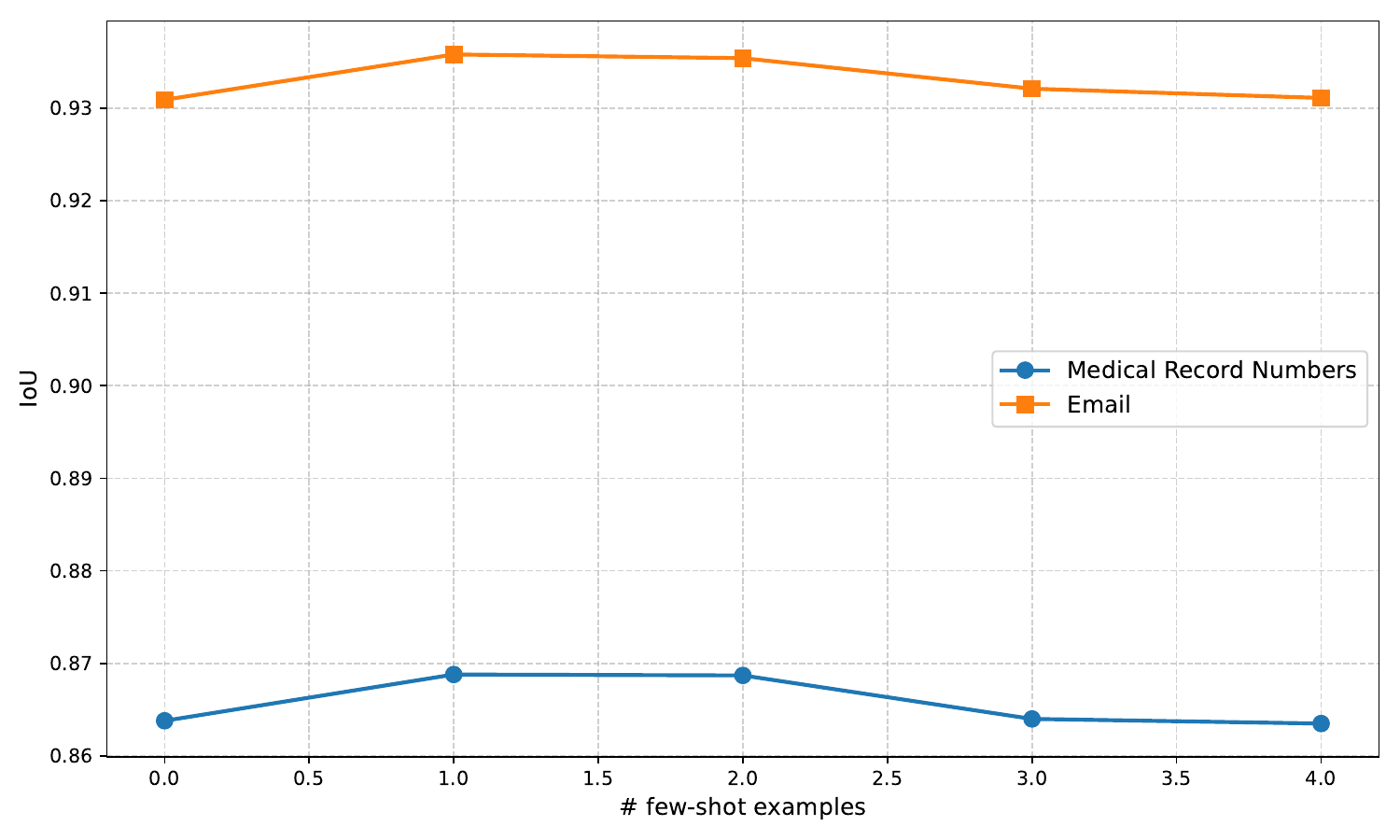}
\caption{IoU performance comparison with different numbers of few-shot examples.}
\label{fig:ab_shot}
\vspace{-2mm}
\end{figure}

For the different number of few-shot examples, we consider using instruction prompts as well as few-shot examples written by human. We focus on the Medical Numbers and Email using CoCo as the base image dataset. And the results are shown in \cref{fig:ab_shot}. We can see that using few-shot examples can boost the performance. However, without using few-shot examples, we can still get a decent result.

\begin{figure}[t!]
\centering
\includegraphics[width=0.48\textwidth]{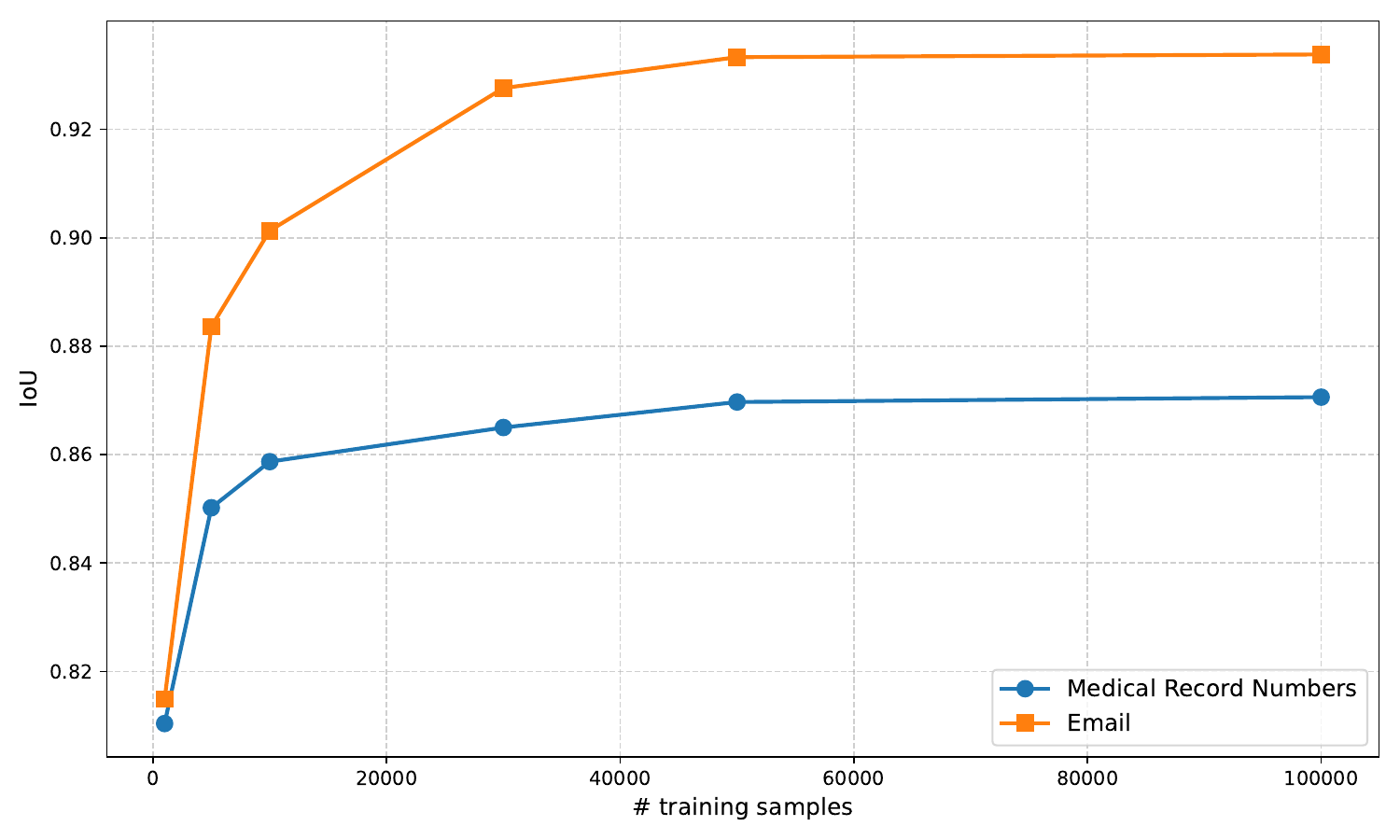}
\caption{IoU performance comparison with different sizes of training dataset}
\label{fig:ab_data}
\end{figure}

In \cref{fig:ab_data}, we present our results for different sizes of training datasets for using CoCo as the base image dataset and instructions from the training set. From the figure, we can observe that using 100k training pairs is more than enough to get a good result, showing the potential ability to use VLMs to de-identify data.

\begin{table*}[ht!]
\centering
\resizebox{\textwidth}{!}{
\begin{tabular}{p{1.5cm}|p{1.5cm}|p{1.5cm}|p{1.5cm}|p{1.5cm}|p{1.5cm}|p{1.5cm}|p{1.5cm}|p{1.5cm}}
\toprule
Model    & Name   & DOB    & SSN    & Email  & Phone Number & Address & Medical Number & Disease Name \\
\midrule
\multicolumn{9}{c}{Evaluation Set Generated by Training Base Image Dataset} \\
\midrule
Full & 0.9478  & 0.9479 & 0.9482 & 0.9482  &  0.9480   &  0.9484  & 0.9478 &   0.9492    \\
Presidio &   0.0007     &  0.0006      & 0.0005     &  0.0006      &  0.0007           &    0.0012     &     0.0004          &     0.0004        \\
\midrule
\multicolumn{9}{c}{Evaluation Set Generated by COCO} \\
\midrule
Full& 0.9470  &  0.9472 &  0.9472  & 0.9472   & 0.9473& 0.9470 & 0.9468  &    0.9467   \\
Presidio & 0.0006 & 0.0005 & 0.0005  &  0.0006    &  0.0006 & 0.0011  &   0.0005 & 0.0004   \\
\midrule
\multicolumn{9}{c}{Evaluation Set Generated by ADE-20K} \\
\midrule
Full     &  0.9196   &   0.9196  &   0.9198     &   0.9198    &   0.9200  &  0.9199   &   0.9197 &  0.9196     \\
Presidio & 0.0002 & 0.0002 & 0.0001  &  0.0002      &  0.0002 & 0.0003  &   0.0001 & 0.0001 \\
\midrule
\multicolumn{9}{c}{Evaluation Set Generated by RITE} \\
\midrule
Full     & 0.9394  &  0.9388   &   0.9398   &    0.9396    & 0.9399    &   0.9397      &   0.9398    &   0.9400     \\
Presidio & 0.0003 & 0.0003 & 0.0003 &  0.0003      &  0.0003 & 0.0007  &   0.0003 & 0.0003  \\
\midrule
\bottomrule
\end{tabular}
}

\caption{Comparative analysis of model performance across information categories, model architectures, and evaluation datasets using mAP as the metric.}
\label{tab:different_image_map}
\end{table*}

\begin{table*}[ht!]
\centering
\resizebox{\textwidth}{!}{
\begin{tabular}{c c c c c c c c c}
\toprule
Model    & Name   & DOB    & SSN    & Email  & Phone Number & Address & Medical Number & Disease Name \\
\midrule
\multicolumn{9}{c}{Instruction Prompts Generated by Gemini1.5} \\
\midrule
Full     &    0.8933   &   0.8932   &    0.8932    &  0.8930  &   0.8931     &  0.8929 &  0.8928 & 0.8933          \\
Presidio &   0.0007     &  0.0006      & 0.0005     &  0.0006      &  0.0007           &    0.0012     &     0.0004          &     0.0004  \\
\midrule
\multicolumn{9}{c}{Instruction Prompts Generated by Human} \\
\midrule
Full     &   0.9221  &  0.9229  & 0.9234 &   0.9224     &   0.9231     &  0.9233    &   0.9223    &   0.9233     \\
Presidio & 0.0006 & 0.0005 & 0.0005  &  0.0006    &  0.0006 & 0.0011  &   0.0005 & 0.0004  \\
\midrule
\bottomrule
\end{tabular}
}
\caption{Performance comparisons for different types of information, different models, and different instruction prompts. The evaluation image set is chosen to evaluation set generated by the training base image dataset using mAP as the metric.}
\label{tab:different_prompts_base_image_map}
\end{table*}

\subsection{Precision and Recall Results}

In the paper, we mainly focus on the F1, which is the balanced metric that considers both precision and recall. To provide a more comprehensive result, we also report precision and recall for the Base Image setting. The results are shown in \cref{tab:precision_recall}.

\begin{table}[h!]
\centering
\begin{tabular}{lcc}
\toprule
\textbf{Information} & \textbf{Precision} & \textbf{Recall} \\
\midrule
Address & 0.9807 & 0.9812 \\
Email & 0.9653 & 0.9786 \\
SSN & 0.9839 & 0.9928 \\
Phone & 0.9621 & 0.9855 \\
DOB & 0.9731 & 0.9969 \\
Med Num & 0.9619 & 0.9910 \\
Name & 0.9629 & 0.9841 \\
Disease & 0.9097 & 0.9427 \\
\bottomrule
\end{tabular}
\caption{Recall and precision of full fine-tuned model with Base Image setting.}
\label{tab:precision_recall}
\end{table}

\subsection{Example on Real-world Dataset}

\begin{figure}[t!]
\centering
\includegraphics[width=0.48\textwidth]{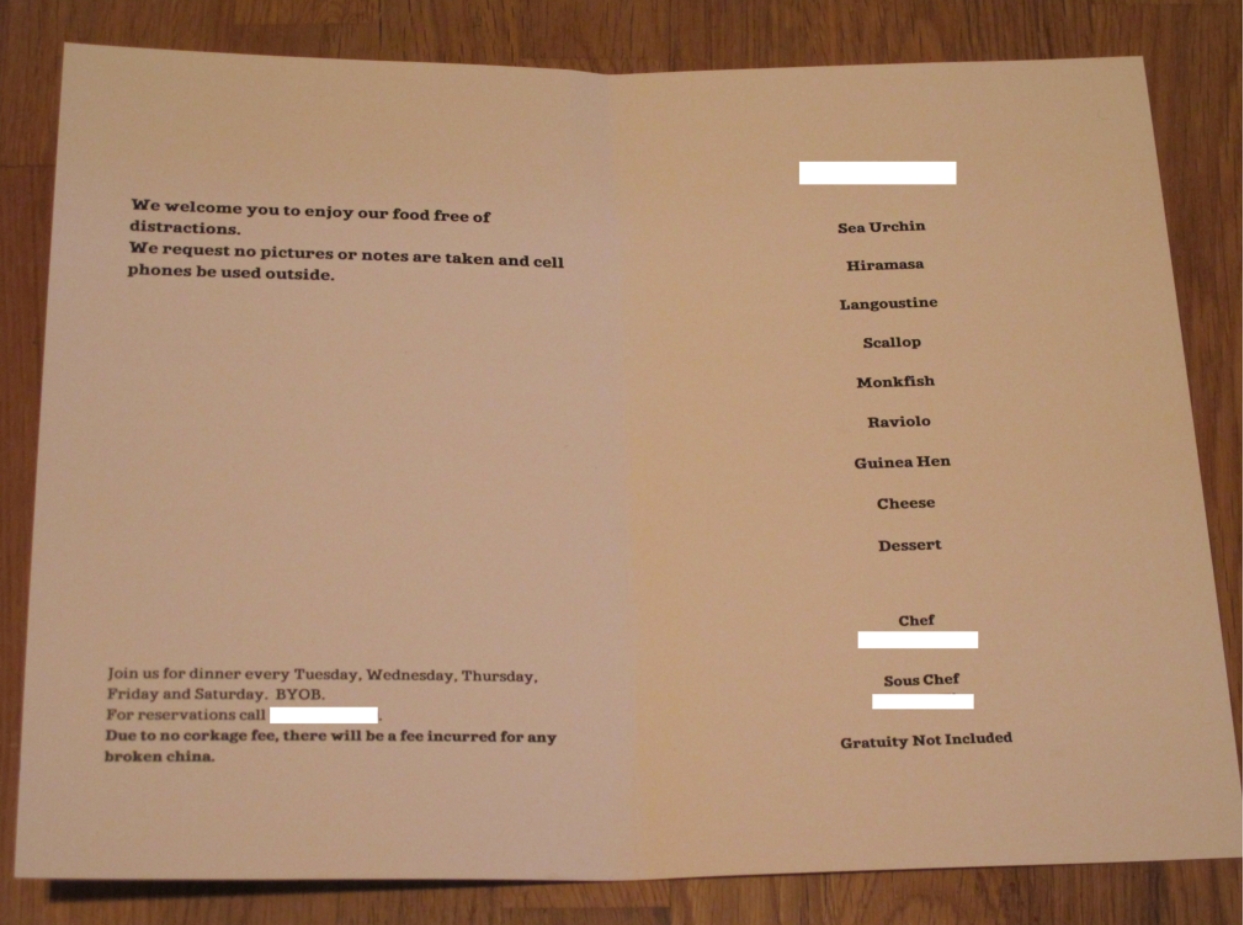}
\caption{A real-world image example that was de-identified by our pipeline.}
\label{fig:real_world}
\end{figure}

In \cref{fig:real_world}, we present an example of applying our fine-tuned model to the real-world dataset. From the figure, we can see that the names and phone numbers are correctly masked by our de-identification pipeline.

\end{document}